\documentclass[conference]{IEEEtran}
\IEEEoverridecommandlockouts
\usepackage{booktabs}
\usepackage{subcaption}
\usepackage{cite}
\usepackage{amsmath,amssymb,amsfonts}
\usepackage{algorithmic}
\usepackage{algorithm}
\usepackage{graphicx}
\usepackage{textcomp}
\usepackage{xcolor}
\usepackage{comment}
\def\BibTeX{{\rm B\kern-.05em{\sc i\kern-.025em b}\kern-.08em
    T\kern-.1667em\lower.7ex\hbox{E}\kern-.125emX}}
\begin{document}

\title{DiSC-Med: Diffusion-based Semantic Communications for Robust Medical Image Transmission
}

\author{\IEEEauthorblockN{Fupei Guo$^*$, Hao Zheng$^*$,  Xiang Zhang$^\star$, Li Chen$^*$, Yue Wang$^\dagger$, Songyang Zhang$^*$}
\IEEEauthorblockA{$^*$University of Louisiana at Lafayette, Lafayette, LA, USA, 70504 \\$^\star$Louisiana State University,  Baton Rouge, LA, USA, 70802 \\$^\dagger$Georgia State University, Atlanta, GA, USA, 30302}
 \IEEEauthorblockA{
\{fupei.guo1, hao.zheng, li.chen,  songyang.zhang\}@louisiana.edu; xzha135@lsu.edu; ywang182@gsu.edu.
 }
}

\maketitle

\begin{abstract}
The rapid development of artificial intelligence has driven smart health with next-generation wireless communication technologies, stimulating exciting applications in remote diagnosis and intervention. To enable a timely and effective response for remote healthcare, efficient transmission of medical data through noisy channels with limited bandwidth emerges as a critical challenge. In this work, we propose a novel diffusion-based semantic communication framework, namely DiSC-Med, for the medical image transmission, where medical-enhanced compression and denoising blocks are developed for bandwidth efficiency and robustness, respectively. Unlike conventional pixel-wise communication framework, our proposed DiSC-Med is able to capture the key semantic information and achieve superior reconstruction performance with ultra-high bandwidth efficiency against noisy channels. 
Extensive experiments on real-world medical datasets validate the effectiveness of our framework, demonstrating its potential for robust and efficient telehealth applications.
\end{abstract}


\section{Introduction}
The development of next-generation wireless communications, such as beyond-fifth-generation (B5G) networking and sixth-generation (6G) technologies, is stimulating many novel applications in daily life, including those in smart health services \cite{10872774}. The advancement in artificial intelligence (AI) and Internet-of-Things (IoT) has enabled remote healthcare and distributed medical data processing as important parts of the future telehealth system \cite{10184989}, which require efficient transmission of medical images. Unlike natural images, medical images, such as computed tomography (CT) and magnetic resonance imaging (MRI) scans, usually have higher dimensions and a wider dynamic range, leading to high storage cost and heavy communication overhead. The demand for timely data transmission makes efficient medical image compression and reconstruction a challenge in telecommunication systems~\cite{9928407}.

Conventional storage and transmission of medical images utilize the DICOM format, which contains both image data, and metadata including patient information and acquisition parameters. Although DICOM supports both lossless and lossy compression, such as JPEG-LS and JPEG, these methods are designed to preserve sufficient information for diagnosis; thus retaining significant redundancy~\cite{review2014compression, review2020versatile}. In remote healthcare scenarios, where a typical communication rate is around 1~Mbps, transmitting a standard CT scan can take 4--7 minutes. This delay becomes critical when handling large volumes of data or time-sensitive diagnostic tasks.

\begin{figure*}[t]
    \centering
    \includegraphics[width=0.88\textwidth]{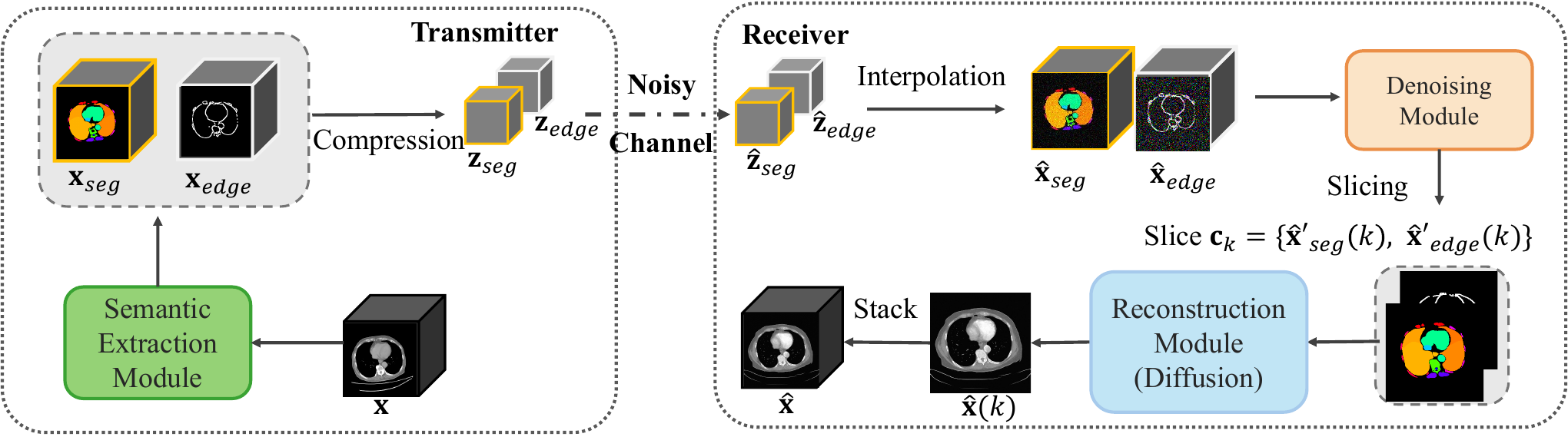}
    \caption{\small
    Overall framework of DiSC-Med: 
    1) \textbf{Transmitter} – The transmitter extracts the semantic representations from the original images, which are compressed to a transmission-friendly low-dimensional representation;
    2) \textbf{Receiver} – The receiver interpolate and denoise the semantic representations, which serve as the conditions of the diffusion models for image reconstruction.}
    \label{fig:backbone}
\end{figure*}

Recent advances in medical image compression focus on learning-based approaches, where neural networks encode the original data into latent representations. One typical method is the variational autoencoder (VAE), which has shown success in the compression of CT and X-ray images ~\cite{liu2022medicalvae, xu2021xrayvae}. In addition, Ballé et al.~\cite{balle2018variational} introduce a variational compression framework with a learned scale hyperprior to model spatial dependencies in the latent space, achieving effective rate–distortion performance. In~\cite{CAEkar2018fully}, a fully convolutional autoencoder (AE) is proposed for high-density mammogram compression, which incorporates arithmetic coding to generate variable-length representations. Other learning-based approaches adopt more advanced architectures, such as vector-quantized VAEs (VQ-VAEs)~\cite{zhou2021vqvae} and Transformers~\cite{tang2023transcs}. Despite some successes, existing works focus on bit-wise data compression and recovery with
performance bounded by Shannon capacity during data transmission \cite{10947303}. However, in real-world applications, downstream tasks often depend on key information, where transmitting the most critical semantic content can suffice for the demand of the receivers. For example, the meniscus plays a more important role in identifying knee injuries{~\cite{meniscus2007}. This motivates goal-oriented semantic communications to convey the key semantics for medical image transmission.

Unlike conventional data-oriented communication framework, goal-oriented semantic communications aim to deliver mission-critical semantic representations to the receiver, rather than reconstructing the entire medical data packets, thereby significantly reducing communication overhead~\cite{9679803}. 
Established examples include joint deep learning enabled semantic communications (DeepSC) \cite{9398576} and masked VQ-VAE \cite{10101778}. With the development of AI technologies, generative learning has attracted significant attention in semantic communications. Utilizing the extracted semantic embedding from the transmitter as constraints, a conditional generative AI (GenAI) model can be deployed to regenerate medical images at the receiver end. This line of research includes generative semantic communications (GESCO) \cite{grassucci2023generative}, diffusion-based goal-oriented communications (Diff-GO) \cite{10615283}, and token communications \cite{qiao2025token}. Despite their promising performance on natural images, semantic communication, as an emerging technology, has been underexplored for medical image transmission. 
To the best of our knowledge, the only known work~\cite{9928407} proposes a domain-enhanced AE using medical images as a case study.
To further facilitate smart healthcare through semantic communications, a critical challenge lies in efficiently integrating conventional medical compression techniques with advanced AI models. 
However, how to handle noise in wireless channels for medical image transmission remains an open problem.

In this work, we introduce a \textbf{Di}ffusion-based \textbf{S}emantic \textbf{C}ommunication for \textbf{Med}ical image transmission, namely \textit{DiSC-Med}, which is based on conditional diffusion models. 
Specifically, segmentation and edge maps are extracted as semantic representations of CT images at the transmitter, which serves as the condition of diffusion model for data regeneration at the receiver. 
To further compress the semantic representation and enhance robustness, specialized sampling and denoising schemes are designed based on deep neural networks. 
The experimental results validate the efficacy of the proposed method in capturing the goal-oriented semantic information and meeting receiver demands under limited and noisy wireless channels.

Our contributions can be summarized as follows:
\begin{itemize}
    \item To the best of our knowledge, this is the first work applying GenAI-based semantic communications in medical image transmission, which has shown a significant improvement in data recovery under limited bandwidth compared to conventional AE-based methods.
    
    \item To balance compression rate and reconstruction quality, the proposed framework leverages the 3D nature of medical images by integrating 3D compression with 2D reconstruction.
    
    \item To enhance the robustness against channel noise, a channel-aware semantic recovery block is introduced for semantic reconstruction and denoising.
    This work considers both continuous and discrete noise models.

    \item Using CT images as an example, our experiments demonstrate superior performance in both semantic characterization and specific downstream tasks. 
\end{itemize}

\section{System Description}
Before presenting the details of DiSC-Med, we first introduce the objective and system model.
\subsection{Objective}
In this work, we focus on the transmission of 3D CT images with anatomical structure segmentation as a downstream task, as shown in Fig. \ref{fig:backbone}.Suppose that $\mathbf{x} \in \mathbb{R}^{D \times H \times W}$ represents a 3D CT volume, where $D$ is the number of axial slices, and $H$ and $W$ denote the height and width of each slice.

Our objective is to encode $\mathbf{x}$ into compressed semantic representations $\mathbf{z}$ at the transmitter, after which the receiver reconstructs the 3D CT image $\hat{\mathbf{x}}$ via diffusion models for anatomical structure segmentation. 
More specifically, we utilize the coarse segmentation map and the edge map as semantic representations to capture the underlying features of CT images. Note that our proposed framework can be easily extended to more general medical data, where metadata, such as patient information and diagnosis records, and additional modalities like k-space representation can be integrated as additional semantic conditions, which we plan to explore in future work.

\subsection{Noise Model} \label{sec: noise}
Denosing the received signals plays an important role in enhancing the final data regeneration and completing the downstream tasks at the receiver. In this work, we consider two noise models: 1) additive white Gaussian noise (AWGN); and 2) discrete bit-wise error.

\subsubsection{AWGN}
In the AWGN model, the transmitted latent representation $\mathbf{z} \in \mathbb{R}^{d}$ is corrupted by additive Gaussian noise during transmission. The received signal $\mathbf{z}'$ is modeled as:
\begin{equation}
    \mathbf{z}' = \mathbf{z} + \mathbf{n}, \quad \mathbf{n} \sim \mathcal{N}(0, \sigma^2 \mathbf{I}),
\end{equation}
where $\mathbf{n}$ is zero-mean Gaussian noise with variance $\sigma^2$. The AWGN can be characterized by the signal-to-noise ratio (SNR) in decibels (dB), which is defined by:
\begin{equation}
\mathrm{SNR}_{\mathrm{dB}} 
= 10 \log_{10} \left( \frac{P_{\text{signal}}}{P_{\text{noise}}} \right),   
\end{equation}
where $P$ denotes the average signal power.

\subsubsection{Bit-wise Error}
We also consider discrete bit-wise or pixel-wise errors for the latent representation.

For the segmentation map, each pixel-wise class label $y_i \in \{0, 1, \dots, C{-}1\}$ is perturbed by a mislabeling process the follows
\(
\mathbb{P}(\hat{y}_i = j \mid y_i = k) = T_{kj}\) with \(j, k \in \{0, 1, \dots, C{-}1\}  
\). $\mathbf{T}=\{T_{kj}\} \in \mathbb{R}^{C \times C}$ is a transition matrix specifying the probability of mislabeling class $k$ as class $j$.

For the binary edge map or latent representation from encoder, i.e., $y_i \in \{0, 1\}$, bit-flip noise is considered with the bit-flip probability defined by
\(
\mathbb{P}(\hat{y}_i = 1 - y_i) = p,
\)
where $\hat{y}_i$ is the received binary value of the mask in the edge map or $\hat{y}_i=\hat{z}_i$ in latent representation.

To quantify the bit-wise corruption, \textit{Bit Error Rate} (BER) is considered, denoted by
\(
\mathrm{BER} = \frac{1}{N} \sum_{i=1}^{N} \mathbb{I}(\hat{y}_i \neq y_i),  
\)
where \( N \) denotes the total number of bits in the semantic representation, and \( \mathbb{I}(\cdot) \) is the indicator function. BER provides a unified metric for assessing the impact of discrete noise on binary and multi-class semantic content.

\section{Method}

The detailed design of DiSC-Med is presented below, with its overall architecture depicted in Fig.~\ref{fig:backbone}.

\subsection{Overall Architecture}

At the transmitter side, a sequence of 3D CT images is first processed by the Semantic Extraction Module, which extracts two types of semantic representations: 1) a segmentation volume, and 2) an edge volume. The segmentation volume provides an abstract representation of anatomical semantics, indicating the spatial distribution of anatomical structures. The edge volume captures finer structural details, offering complementary and texture information for reconstruction.
To reduce communication overhead, both semantic volumes are compressed into compact 3D latent representations using a learned compression module to be introduced in Section \ref{sec:compress}, after which the latent representation is transmitted to the receiver through a noisy communication channel.

At the receiver side, the semantic latent representations are passed through an interpolation module and a denoising module, which recover the 3D segmentation and edge volumes. To reduce computational complexity, we decompose the 3D semantic representations into a set of 2D slices, which are fed into a pre-trained conditional diffusion model for 2D CT image reconstruction. Finally, all 2D predictions are reassembled into a 3D volume for downstream tasks, such as anatomical structure segmentation task in this work.

\subsection{Transmitter}
\subsubsection{Semantic Extraction Module}
As aforementioned, segmentation and edge volumes are extracted as semantic representations for CT images. In this work, a pre-trained 2D semantic segmentation model $f_{\text{seg}}$ is applied slice-by-slice to the 3D CT image, generating the segmentation maps $\mathbf{x}_{\text{seg}} = f_{\text{seg}}(\mathbf{x})\in \mathbb{R}^{D \times H \times W}$.
Specifically, a medical foundation model, TotalSegmentator~\cite{wasserthal2023totalsegmentator}, is used for semantic segmentation.
Each obtained slice contains multi-channel semantic logits or can be further coded as one-hot representations corresponding to anatomical structures. 

Similarly, a 2D edge detection model $f_{\text{edge}}$ is designed to extract boundary information and texture details for each slice, resulting in a stacked edge volume
$\mathbf{x}_{\text{edge}} = f_{\text{edge}}(\mathbf{x})
\in \mathbb{R}^{D \times H \times W}$. 
Specifically, we first apply another foundation model, MedSAM~\cite{ma2023medsam}, to obtain diversified coarse anatomical masks, followed by the Canny edge detector to compute structural boundaries.
The edge information complements the segmentation map by preserving fine-grained spatial details.


\subsubsection{Compression} \label{sec:compress}

After obtaining the segmentation volume $\mathbf{x}_{\text{seg}}$ and edge volume $\mathbf{x}_{\text{edge}}$ from the Semantic Extraction Module, we apply a compression step to further reduce data redundancy prior to transmission. 
In particular, we downsample $\mathbf{x}_{\text{seg}}$ and $\mathbf{x}_{\text{edge}}$ via strided sampling along the channel and spatial dimensions, where we retain every second channel and every fourth pixel in height and width, resulting in compressed representations $\mathbf{z}_{\text{seg}}$ and $\mathbf{z}_{\text{edge}}$ with dimensions $(D/2, H/4, W/4)$.


\subsection{Receiver}
\subsubsection{Interpolation Module}
With the received  latent representations $\hat{\mathbf{z}}_{\text{seg}}$ and $\hat{\mathbf{z}}_{\text{edge}}$, we first restore them into the original spatial resolution via trilinear interpolation, i.e.,
\(
\hat{\mathbf{x}}_{\text{seg}} = \text{Tri}(\hat{\mathbf{z}}_{\text{seg}})\) and
\(\hat{\mathbf{x}}_{\text{edge}} = \text{Tri}(\hat{\mathbf{z}}_{\text{edge}})
\), where $\text{Tri}(\cdot)$ denotes the trilinear interpolation operation. In particular, the interpolation is applied independently to each channel of the latent tensors, reconstructing the feature maps from $(D/2, H/4, W/4)$ back to $(D, H, W)$.
For each interpolated voxel at position $(d, h, w)$, the value is computed as a weighted sum of the eight nearest neighbors:
\begin{equation}
\hat{v}(d, h, w) = \sum_{i=0}^{1}\sum_{j=0}^{1}\sum_{k=0}^{1} w_{ijk} \cdot v(d_i, h_j, w_k),
\end{equation}
where $w_{ijk}$ are interpolation weights determined by the relative distance between the point $(d, h, w)$ and its eight neighboring grid points.

\begin{figure}[t]
    \vspace*{1mm}
    \centering
    \includegraphics[width=1\linewidth]{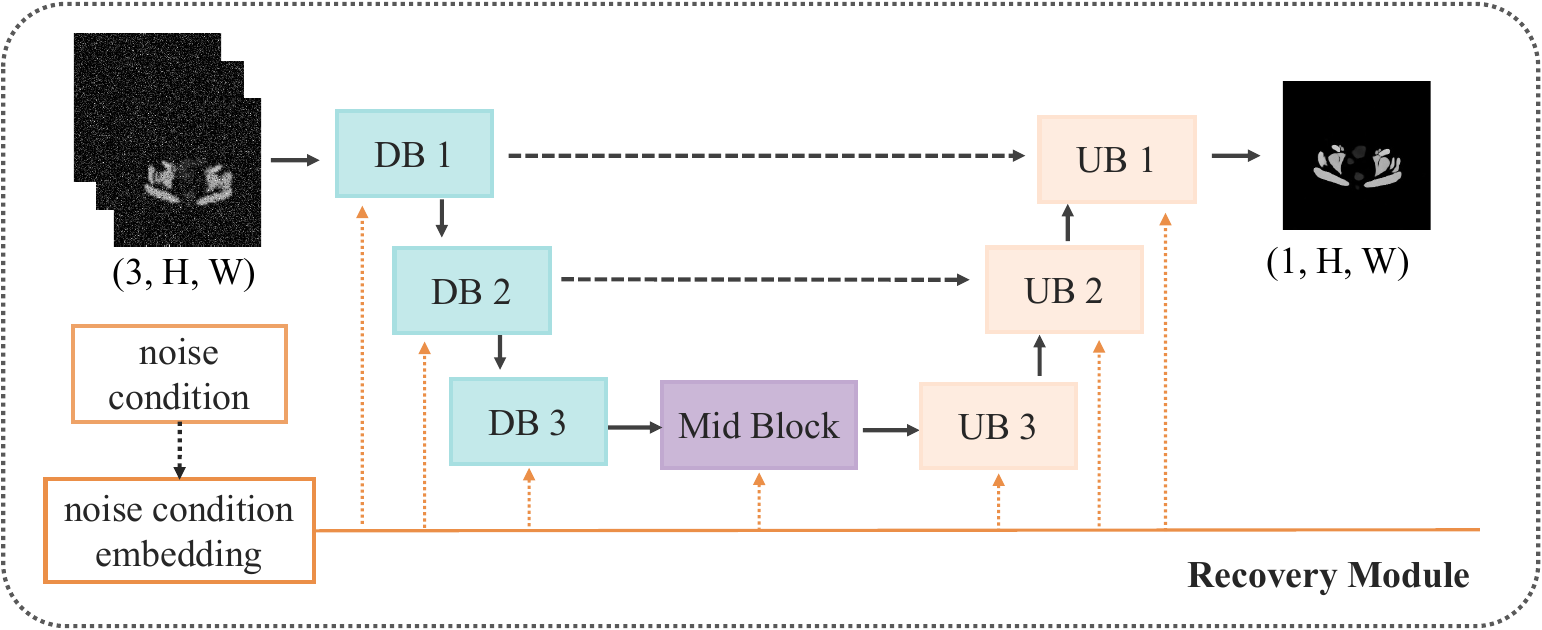} 
    \caption{Structure of the Denoising Module: DB stands for DownBlock and UB stands for UpBlock.}
    \label{fig:recovery_module}
\end{figure}

\subsubsection{Denoising Module}
With the interpolated semantic representations, a channel-aware denoising module is designed, which aims to address the impact from two types of noise: i) channel noise, and ii) errors from interpolation.


Instead of directly applying a computationally intensive 3D neural network model, we leverage the strong inter-slice dependencies of anatomical structures by employing a slice-wise 2D U-Net as shown in Fig. \ref{fig:recovery_module}. Specifically, we decompose the interpolated latent volume along the axial axis into individual slices. To recover each target slice, we construct a 3-channel input by stacking the target slice and its two adjacent slices along the channel dimension, which is then fed into the semantic denoising (recovery) neural network. To address channel noises, we adopt a channel-aware \textit{Conditional U-Net} architecture, where channel noise information, such as SNR and BER, is injected into the network to guide the recovery process. Specifically, we first encode the current channel state information into a learnable embedding vector and then expand it to match the dimensions of each intermediate feature map.

During inference, the noise condition $\mathbf{c}$ is integrated into the U-Net via feature-wise affine transformations at designated layers, allowing the network to dynamically adapt its denoising and semantic restoration behavior according to channel quality. With this module, we recover a better semantic representation with denoising and compensation, that is, $\textbf{c}=\{\hat{\mathbf{x}}'_{seg}, \hat{\mathbf{x}}'_{edge}\}$, where $\hat{\mathbf{x}}'_{seg}$ and $\hat{\mathbf{x}}'_{edge}$ are the denoised segmentation and edge volumes.

\begin{figure}[t]
    \centering
    \includegraphics[width=1\linewidth]{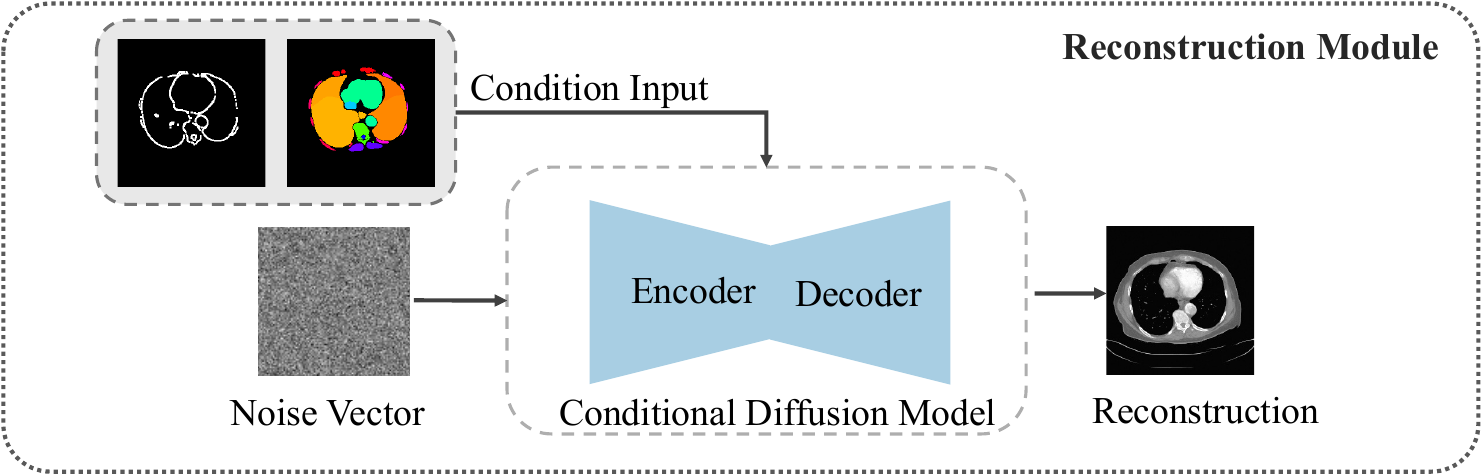} 
    \vspace{-2mm}  %
    \caption{Structure of the reconstruction module.}
    \label{fig:reconstruct_module}
    \vspace{-2mm}  %
\end{figure}

\subsubsection{Reconstruction Module}
With the recover semantic representation, we adopt a conditional Denoising Diffusion Probabilistic Model (DDPM) \cite{ho2020denoising} as the final stage of CT image reconstruction, as illustrated in Fig.~\ref{fig:reconstruct_module}.

The restored 3D semantic volumes are converted into 2D slice-wise maps, i.e., $\hat{\mathbf{x}}'_{\text{seg}}(k)$ and $\hat{\mathbf{x}}'_{\text{edge}}(k)$ for the $k$th slice, which are concatenated to form the conditioning input: 
\begin{equation}
    \mathbf{c}_k = \text{Concat}(\hat{\mathbf{x}}'_{\text{seg}}(k), \hat{\mathbf{x}}'_{\text{edge}}(k)).
\end{equation}
The DDPM learns to generate high-quality images conditioned on $\mathbf{c}_k$ through a reverse denoising process. Starting from Gaussian noise $\mathbf{x}_T(k) \sim \mathcal{N}(0, \mathbf{I})$, the model iteratively denoises the sample through a learned noise estimation network:
\begin{equation}\small
    \mathbf{x}_{t-1}(k) = \frac{1}{\sqrt{\alpha_t}}\left( \mathbf{x}_t(k) - \frac{1 - \alpha_t}{\sqrt{1 - \bar{\alpha}_t}} \epsilon_\theta(\mathbf{x}_t(k), t, \mathbf{c}_k) \right) + \sigma_t \mathbf{z},
\end{equation}
where $\alpha_t$ is the noise schedule, $\bar{\alpha}_t = \prod_{s=1}^{t} \alpha_s$, and $\mathbf{z} \sim \mathcal{N}(0, \mathbf{I})$ is standard Gaussian noise. The denoising steps are performed for $t = T, \ldots, 1$, eventually yielding the final reconstructed slice $\hat{\mathbf{x}} = \mathbf{x}_0$.
Finally, all reconstructed 2D slices $\{\hat{\mathbf{x}}(k)\}_{i=1}^D$ are reassembled into a 3D CT volume, which can be used for downstream tasks such as anatomical structure segmentation and disease diagnosis. 
In this work, the diffusion model adopts a U-Net backbone with five levels of downsampling and upsampling, where the number of channels ranges from 256 to 1024. To improve spatial feature modeling, attention mechanisms are incorporated at the lowest-resolution stages, including the bottleneck layer, during the reconstruction process.

\section{Experiments}

\subsection{Dataset}

We used a subset of abdominal CT of 16 patients derived from the AMOS dataset~\cite{ji2022amos}, each containing approximately 100 to 400 axial slices. The original CT slices, with a resolution of $512 \times 512$ pixels, were resized to $256 \times 256$ to reduce computational complexity. For the DDPM-based reconstruction module, CT intensities were clipped in range of $[-400, 400]$ in terms of Hounsfield units (HU) and then normalized to $[0, 1]$. The dataset was split into 10 patients for training and 6 patients for testing. A separate set of 186 3D CT volumes was used to train the Denoising Module, enabling it to learn the spatial and contextual features necessary for volumetric semantic restoration. Note that although scans from only 16 patients were considered, the dataset comprised 29,184 $2D$ images, which was sufficient to support robust model training and evaluate data transmission performance.

\subsection{Overall Performance Without Noise}
To validate the performance, we compared the proposed method with several representative compression baselines in both semantic communications and medical image compression methods, including deep convolutional autoencoder (DCAE) \cite{fettah_convolutional_2024}, ResVAE that incorporates residual connections into VAE \cite{liu2022medicalvae}, hierarchical vector-quantized VAE (VQ-VAE-2) \cite{razavi2019generating}, and compressive autoencoder (CAE) that enhances spatial coherence by capturing structural information through learned transformations \cite{CAEkar2018fully}.

\begin{table}[t]
\vspace{1mm}
\centering
\caption{Comparison with baselines in clean data.}
\label{tab:baseline_compression}
\begin{tabular}{lccc}
\toprule
\textbf{Method} & \textbf{LPIPS} $\downarrow$ & \textbf{FID} $\downarrow$ & \textbf{CR} $\uparrow$ \\
\midrule
DCAE\cite{AEhinton2006reducing}        & 0.1631 & 186.82  & 51.2x \\
ResVAE\cite{vaekingma2013auto}       & 0.1604 & 170.78 & 51.2x \\
VQ-VAE-2\cite{vqvaevan2017neural}    & 0.1218 & 130.07 & 51.2× \\
CAE\cite{CAEkar2018fully}         & 0.1625 & 166.73 & 42.7x \\
DiSC-Med      & 0.1499 & 100.91 & 50.69× \\
\bottomrule
\end{tabular}

\end{table}
\begin{figure}[t]
    \centering
    \includegraphics[width=0.95\linewidth]{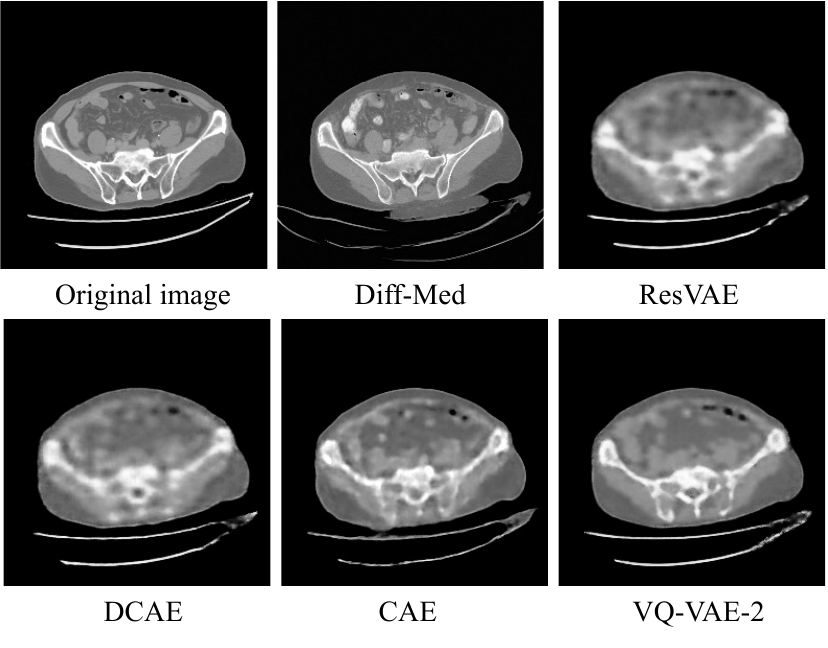} 
    \caption{Visualization of reconstructed CT images.}
    \label{fig:recon_visual}
    \vspace{-3mm}
\end{figure}

\begin{table*}[ht]
\centering
\vspace{1mm}
\caption{Performance of anatomical structure segmentation under noiseless transmission across different methods. D: Dice coefficient; I: weighted mean Intersection over Union; H: 95th
percentile Hausdorff Distance. The best scores are in \textbf{bold}.}
\label{tab:downstream_metrics}
\resizebox{\textwidth}{!}{%
\begin{tabular}{l|ccc|ccc|ccc|ccc|ccc|ccc}
\toprule
\textbf{Method} 
& \multicolumn{3}{c|}{\textbf{Spleen }} 
& \multicolumn{3}{c|}{\textbf{Right Kidney}} 
& \multicolumn{3}{c|}{\textbf{Left Kidney }} 
& \multicolumn{3}{c|}{\textbf{Liver}} 
& \multicolumn{3}{c|}{\textbf{Aorta }} 
& \multicolumn{3}{c}{\textbf{Postcava}} \\
\midrule
& D & I & H & D & I & H & D & I & H & D & I & H & D & I & H & D & I & H \\
\midrule
DCAE       
& 0.8707 & 0.7711 & 2.45 
& 0.8149 & 0.6877 & 3.61 
& 0.8412 & 0.7259 & 3.16 
& 0.9111 & 0.8368 & 3.00 
& 0.6437 & 0.4746 & 7.28 
& 0.6290 & 0.4588 & 7.55 \\
CAE      
& 0.8909 & 0.8032 & 2.00 
& 0.8514 & 0.7412 & 3.16 
& 0.8565 & 0.7491 & 3.16 
& 0.9013 & 0.8204 & 3.74 
& 0.7574 & 0.6096 & 4.24 
& 0.6803 & 0.5154 & 5.10 \\
ResVAE      
& 0.8411 & 0.7258 & 3.00 
& 0.7893 & 0.6519 & 4.12 
& 0.8304 & 0.7100 & 3.16 
& 0.8972 & 0.8136 & 3.61 
& 0.6131 & 0.4420 & 8.05 
& 0.7000 & 0.5385 & 5.48 \\

VQ-VAE-2   
& \textbf{0.9119} & \textbf{0.8381} & \textbf{1.41} 
& 0.8616 & 0.7568 & \textbf{3.61} 
& 0.8902 & 0.8021 & \textbf{3.00} 
& 0.9118 & 0.8379 & \textbf{2.45} 
& 0.7918 & 0.6554 & 4.00 
& 0.6337 & 0.4638 & 6.08 \\
DiSC-Med      
& 0.7735 & 0.6306 & 5.66 
& \textbf{0.9258} & \textbf{0.8618} & 1.41 
& \textbf{0.8990} & \textbf{0.8165} & 3.16 
& \textbf{0.9124} & \textbf{0.8388} & 5.39 
& \textbf{0.9153} & \textbf{0.8438} & \textbf{2.00} 
& \textbf{0.8190} & \textbf{0.6935} & \textbf{2.24} \\
\midrule
\textbf{Method} 
& \multicolumn{3}{c|}{\textbf{Pancreas }} 
& \multicolumn{3}{c|}{\textbf{Right Adrenal Gland }} 
& \multicolumn{3}{c|}{\textbf{Left Adrenal Gland }} 
& \multicolumn{3}{c|}{\textbf{Duodenum }} 
& \multicolumn{3}{c|}{\textbf{Prostate/Uterus }} 
& \multicolumn{3}{c}{\textbf{Average}} \\
\midrule
& D & I & H & D & I & H & D & I & H & D & I & H & D & I & H & D & I & H \\
\midrule
DCAE       
& 0.6443 & 0.4752 & 10.91 
& 0.1301 & 0.0696 & 7.11 
& Fail     & Fail     & Fail 
& 0.4364 & 0.2791 & 12.17 
& 0.6546 & 0.4866 & 7.00
&  0.7851     & 0.6592     & 4.51 \\
CAE      
& 0.6158 & 0.4449 & 4.58 
& 0.2191 & 0.1230 & 6.95 
& Fail     & Fail     & Fail 
& 0.5490 & 0.3784 & 10.34 
& 0.5413 & 0.3711 & 8.06 
& 0.8230     & 0.7065     &  3.57  \\
ResVAE     
& 0.5706 & 0.3992 & 23.72 
& 0.1186 & 0.0630 & 13.07 
& Fail     & Fail     & Fail 
& 0.4377 & 0.2801 & 35.01 
& 0.5505 & 0.3798 & 10.05
& 0.7785     & 0.6470     & 4.57  \\
VQ-VAE-2    
& 0.6585 & 0.4908 & 7.48 
& 0.1938 & 0.1073 & 12.21 
& 0.0767 & 0.0399 & 13.91 
& 0.5847 & 0.4132 & \textbf{6.71} 
& 0.6673 & 0.5008 & 5.92
& 0.8335 & 0.7257 & 3.42  \\
DiSC-Med     
& \textbf{0.7730} & \textbf{0.6301} & \textbf{3.00} 
& \textbf{0.6755} & \textbf{0.5100} & \textbf{2.24} 
& \textbf{0.4114} & \textbf{0.2590} & \textbf{8.15} 
& \textbf{0.6367} & \textbf{0.4670} & 11.58 
& \textbf{0.6370} & \textbf{0.4674} & \textbf{5.39}
& \textbf{0.8742}  &  \textbf{0.7808} & \textbf{3.31}  \\
\bottomrule
\end{tabular}%
}
\end{table*}

We evaluated the performance with two standard perceptual and distortion-based metrics: the Learned Perceptual Image Patch Similarity (LPIPS) and the Fréchet Inception Distance (FID). 
For fair comparison, we carefully tuned the hyperparameters of each baseline method to ensure that all methods were under a similar compression ratio (CR). The quantitative results were presented in Table \ref{tab:baseline_compression}, where our proposed method consistently outperformed the baselines. Qualitative results in Fig. \ref{fig:recon_visual} further demonstrated that our method was able to reconstruct finer details and achieve high-fidelity data regeneration.

\subsection{Performance on Downstream Tasks}
To further evaluate the performance of data reconstruction at the receiver end, we tested the performance of regenerated CT images in anatomical structure segmentation. 
Specifically, we employed a pre-trained anatomical structure segmentation model~\cite{chen2024versatile} as a blind test, ensuring that the model had no prior exposure to the reconstructed data.

\subsubsection{Evaluation Metric}
To evaluate the performance, we adopted three widely used metrics: Dice coefficient (Dice), weighted mean Intersection over Union (mIoU\textsubscript{w}), and 95th percentile Hausdorff Distance (HD95). 
These metrics assess segmentation quality from multiple perspectives, including region overlap, class balance, and boundary accuracy. Let there be $C$ foreground classes. For each class $c$, let $\mathrm{TP}_c$, $\mathrm{FP}_c$, and $\mathrm{FN}_c$ denote the number of pixels that are true positives, false positives, and false negatives, respectively.

The \textbf{Dice coefficient} for class $c$ is computed by:
\begin{equation}
\mathrm{Dice}_c = \frac{2\mathrm{TP}_c}{2\mathrm{TP}_c + \mathrm{FP}_c + \mathrm{FN}_c},
\end{equation}

The \textbf{IoU} for each class is computed by:
\begin{equation}
\mathrm{IoU}_c = \frac{\mathrm{TP}_c}{\mathrm{TP}_c + \mathrm{FP}_c + \mathrm{FN}_c},
\end{equation}

\textbf{HD95} measures the 95th percentile of the Hausdorff Distance between predicted and ground-truth boundaries, capturing worst-case deviation while remaining robust to outliers.

\subsubsection{Performance}
The anatomical structure segmentation performance was summarized in Table \ref{tab:downstream_metrics}. As the results showed, our proposed method achieved superior performance across most evaluation metrics for all anatomical structure. 
In particular, for smaller structures such as the adrenal gland, conventional AE-based approaches often performed poorly and occasionally failed to detect them. In contrast, our DiSC-Med reliably captured their features, demonstrating the effectiveness of the proposed framework in goal-oriented semantic communication for medical image transmission.

\subsection{Performance under Noisy Channels}
We presented the results under noisy channel conditions. 
As described in Section~\ref{sec: noise}, We considered two noisy models: AWGN noise and bit-wise error. 
The reconstruction quality of CT images and the performance on downstream tasks under varying noise levels were shown in Fig. \ref{fig:snr_curves} and Fig. \ref{fig:ber_curves} for AWGN and bit-wise error, respectively. 
Benefiting from our channel-aware denoising module, which compensated for interpolation errors and mitigated channel noise, our DiSC-Med consistently achieved superior performance in both FID score and downstream task accuracy under different noise levels, demonstrating the robustness and efficacy of the framework.

\begin{figure}[t]
    \centering
    \begin{subfigure}[b]{0.5\linewidth}
        \centering
        \includegraphics[width=\linewidth]{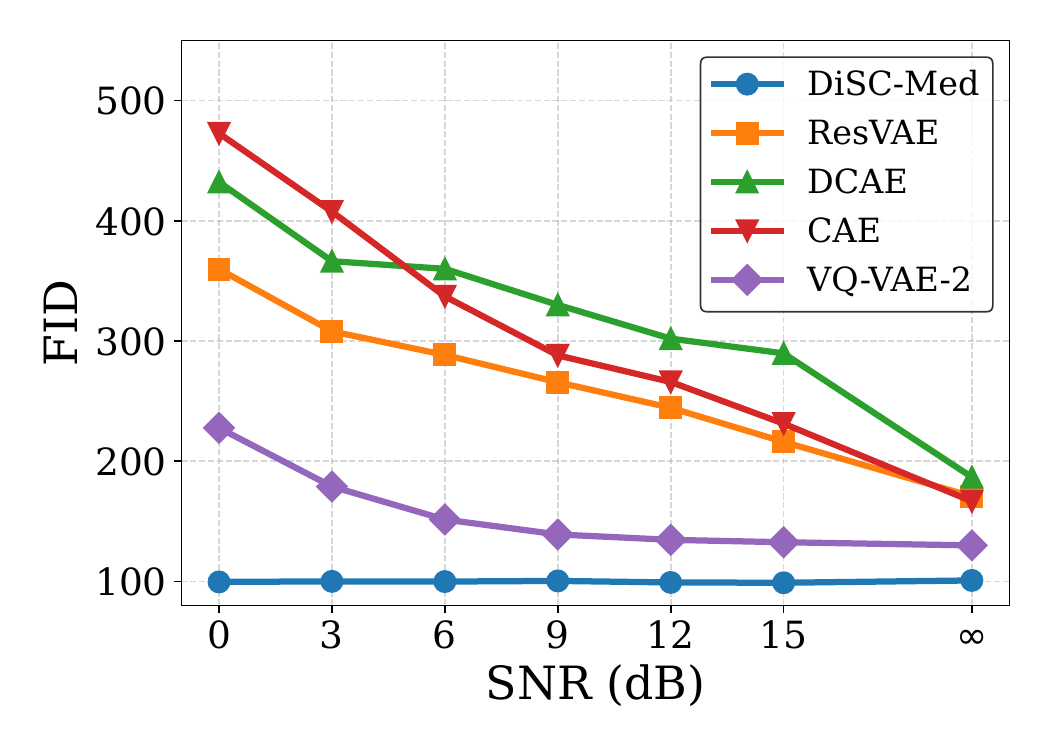}
        \label{fig:fid_snr}
    \end{subfigure}
    \hspace{-0.04\linewidth} 
    \begin{subfigure}[b]{0.5\linewidth}
        \centering
        \includegraphics[width=\linewidth]{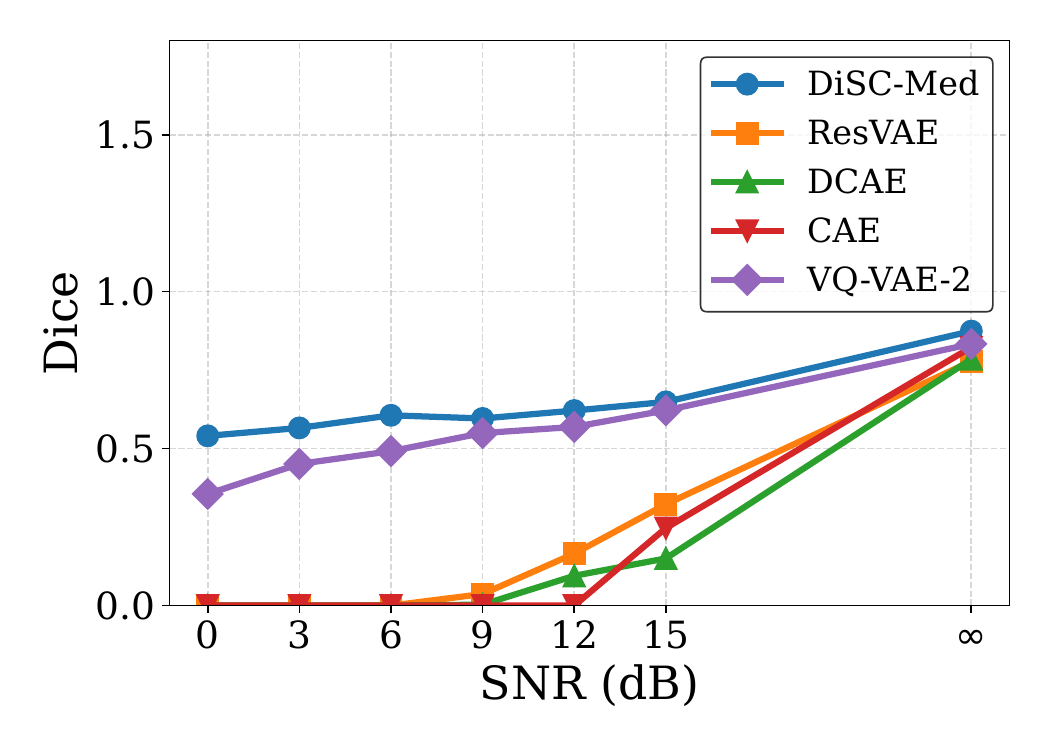}
        \label{fig:dice_snr}
    \end{subfigure}
    \vspace{-4mm}  
    \caption{Performance under AWGN channels: FID $\downarrow$ vs. SNR (left) and Dice $\uparrow$ vs. SNR (right).}

    \vspace{-2mm}
    \label{fig:snr_curves}
\end{figure}

\begin{figure}[t]
    \centering
    \begin{subfigure}[b]{0.51\linewidth}
        \centering
        \includegraphics[width=\linewidth]{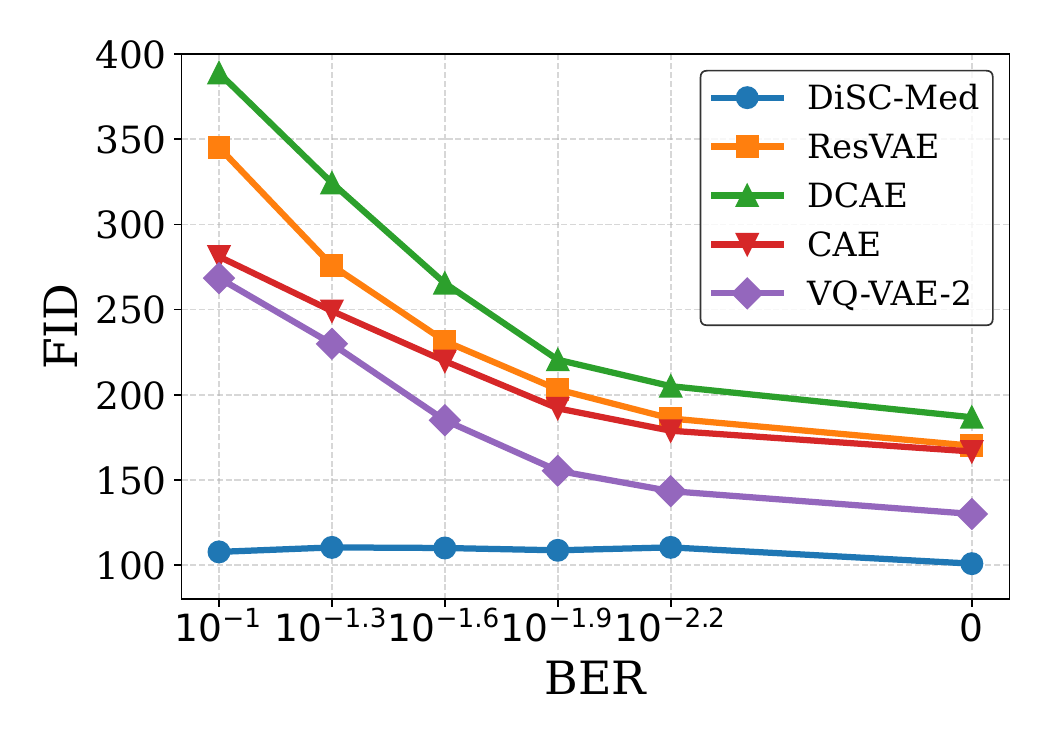}
        \label{fig:fid_ber}
    \end{subfigure}
    \hspace{-0.04\linewidth} 
    \begin{subfigure}[b]{0.5\linewidth}
        \centering
        \includegraphics[width=\linewidth]{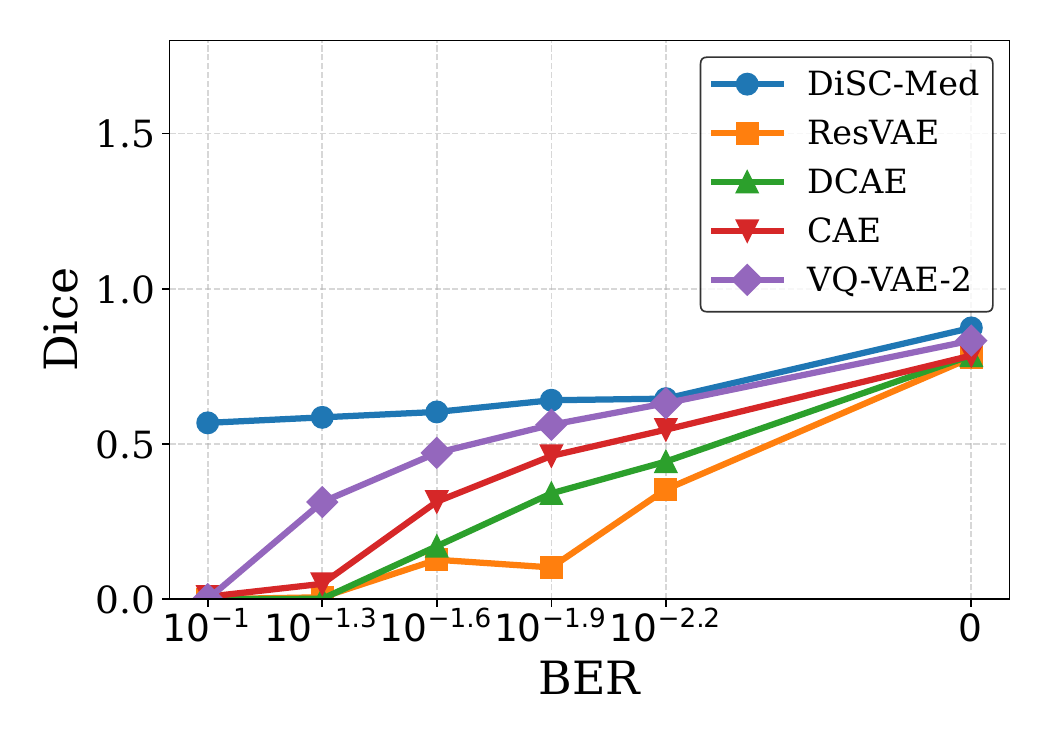}
        \label{fig:dice_ber}
    \end{subfigure}
    \vspace{-9mm}  
    \caption{Performance under bit-wise errors: FID $\downarrow$ vs. BER (left) and Dice $\uparrow$ vs. BER (right).}
    \vspace{-3mm}
    \label{fig:ber_curves}
\end{figure}

\section{Conclusion}

In this work, we proposed a semantic communication framework, DiSC-Med, for efficient medical image transmission and reconstruction using diffusion models. 
By leveraging semantic embeddings, our approach significantly reduces communication overhead while preserving essential structural and semantic information in the original medical images. 
To mitigate channel noise, we introduced a channel-aware denoising module to enhance reconstruction quality at the receiver end. 
Experimental results on CT images with anatomical structure segmentation as a downstream tasks under noise validated the effectiveness of DiSC-Med in efficient and robust medical image transmission. 

\bibliographystyle{IEEEtran} 
\bibliography{reference}

\end{document}